\documentclass[sigconf,screen,nonacm]{acmart}

\setcopyright{none}
\renewcommand\footnotetextcopyrightpermission[1]{}
\usepackage{booktabs}
\usepackage{multirow}
\usepackage{array}
\usepackage{amsmath}
\usepackage{bm}
\usepackage{mathtools}
\usepackage{enumitem}

\newcommand{\dataset}{\mathcal{D}}
\newcommand{\refdata}{\mathcal{R}}
\newcommand{\utility}{\mathbf{u}}
\newcommand{\realism}{\mathbf{r}}
\newcommand{\sourcegap}{\mathbf{s}}

\title{Improving the Realism of Synthetic Clinical Benchmarks Under Utility Constraints}

\author{Omid Bazgir \textsuperscript{*} , Md Nasir, Jacob Hoffman, Yang Yang, Manu Agrawal, Anusua Trivedi, Vinay Rao Dandin, Chris Gibbons, Christine Swisher}
\affiliation{%
  \institution{Oracle Health and Life Sciences}
  \country{United States}
}
\email{omid.bazgir@oracle.com}

\begin{abstract}
Synthetic clinical benchmarks for enterprise AI agents can pass existing utility checks and still remain structurally unrealistic, especially in privacy-sensitive healthcare settings where operational data are hard to access. We study how to improve such benchmarks without breaking the downstream utility checks already used in practice.

We formulate benchmark revision as \emph{utility-constrained realism improvement}: dataset changes should increase realism while staying above an operational utility floor. We instantiate this idea on a care-gap benchmark derived from Synthea-generated patients exercised through demonstration electronic health record workflows and then processed by the same downstream pipeline as operational data. Realism is measured through missingness structure, simplicity, structural plausibility, and population alignment.

The baseline benchmark is extremely thin: sampled-pair missingness is $79.44\%$, only $12.75\%$ of rows are actionable, $38.94\%$ of patients have zero actionable measures, and top-three token concentration reaches $100.0\%$. Two deterministic revisions improve these panels while remaining above the current utility floor, whereas a naive densification control preserves unrealistic templating. We further show that internal benchmark realism and source fidelity to an aggregate operational reference are related but distinct objectives. These results suggest that synthetic benchmark quality should be optimized explicitly, with utility treated as one constraint rather than as sufficient evidence of realism.
\end{abstract}


\begin{document}
\maketitle

\begingroup
\renewcommand{\thefootnote}{\fnsymbol{footnote}}
\footnotetext[1]{Corresponding author: Omid Bazgir (omid.bazgir@oracle.com)}
\endgroup

\section{Introduction}
Current utility checks can fail to detect unrealistic synthetic benchmarks. In our setting, evaluator-derived utility metrics continued to pass even when the benchmark was extremely sparse, heavily templated, and dominated by patient rows with weak evidence context. That is a serious benchmarking failure mode for enterprise healthcare agents: if utility metrics are partially invariant to realism defects, then a dataset can look operationally acceptable while still failing to represent the difficulty and ambiguity of deployment data.

Enterprise AI agents in healthcare are increasingly expected to reason over longitudinal patient records, apply eligibility logic, and generate patient- or operator-facing recommendations in environments where operational data are highly sensitive. Synthetic data and demonstration environments are therefore attractive because they reduce governance friction and enable rapid iteration. However, these advantages create a benchmark-quality problem: a dataset can be easy to use, operationally compatible, and even downstream-useful while still being unrealistically sparse, templated, or clinically narrow.

That failure mode matters at the agent level. In enterprise healthcare settings, many agent pipelines rely on retrieved evidence snippets and bounded context windows to support multi-step reasoning~\cite{lewis2020rag,liu2024lost}. If many rows are non-actionable, retrieval returns less patient-specific evidence; if temporal fields are missing, recency and due-date reasoning become harder to ground; and if descriptions are over-templated, retrieval signal becomes less discriminative. These conditions can plausibly increase generic or weakly grounded responses, even when coarse evaluator scores remain acceptable.

This paper studies that problem for a care-gap benchmark supporting enterprise agent tasks such as patient-level gap interpretation, evidence summarization, and outreach generation. The benchmark originates from Synthea-generated synthetic patients~\cite{walonoski2018synthea} that were exercised through demonstration electronic health record workflows and then processed by the same downstream pipeline as operational client data. That provenance preserves realistic pipeline semantics, but it can still suppress the heterogeneity, missingness, and ambiguity seen in naturally occurring clinical populations.

Our contribution is mainly methodological. We propose \emph{utility-constrained realism improvement}: benchmark revisions should improve realism while staying above the operational utility floor already used by the system. The key idea is to treat utility as a guardrail rather than as sufficient evidence of benchmark quality. The healthcare benchmark in this paper serves as a concrete testbed for that refinement recipe.

This framing builds on prior work showing that synthetic-health-data quality is inherently multidimensional~\cite{giuffre2023synthetic,kaabachi2025scoping}; that synthetic EHR generation ranges from scenario engines such as Synthea~\cite{walonoski2018synthea} to learned record generators~\cite{choi2017medgan,li2023mixedtype,yoon2023ehrsafe}; and that benchmarking, replicability, and drift analyses expose different failure modes~\cite{zhang2022benchmarking,tucker2020fidelity,elemam2024replicability,zhang2022drift}. We use those observations to motivate a practical refinement recipe: improve realism under an explicit utility constraint, then evaluate source fidelity separately from internal realism.

\section{Related Work on Synthetic Data Improvement}
Prior work on synthetic health data has largely emphasized privacy, statistical resemblance, and downstream-task transfer rather than the narrower question studied here: how to improve a benchmark that is already operationally usable but structurally too simple. Reviews argue that fidelity must be assessed through multiple criteria~\cite{giuffre2023synthetic,kaabachi2025scoping}; generation papers span scenario engines such as Synthea~\cite{walonoski2018synthea} and learned EHR generators~\cite{choi2017medgan,li2023mixedtype,yoon2023ehrsafe}; and evaluation studies show that benchmarking, software testing, replicability, and drift control expose different weaknesses~\cite{zhang2022benchmarking,tucker2020fidelity,elemam2024replicability,zhang2022drift}. Our setting is closer to dataset refinement than to end-to-end generation: the benchmark preserves pipeline semantics, but its scenario-driven provenance can still suppress heterogeneity, ambiguity, and realistic missingness. We therefore treat realism as an explicit objective rather than as a post hoc similarity score.

\section{Improvement Objective}
Let $\dataset = \{x_i\}_{i=1}^n$ be a benchmark dataset for a clinical pipeline and let $\utility(\dataset) \in \mathbb{R}^{m}$ denote the vector of task-utility metrics currently used by the pipeline. In our case, these include evaluator-derived measures such as coverage and safety for benchmark subtasks. Let $\realism(\dataset) \in \mathbb{R}^{p}$ denote a vector of realism metrics, partitioned into panels such as missingness structure, language concentration, structural plausibility, and population alignment.

We treat benchmark revision as a constrained optimization problem:
\begin{equation}
\begin{aligned}
\max_{\dataset'} \quad & \sum_{k=1}^{p} w_k \Delta r_k(\dataset', \dataset) \\
\text{subject to} \quad & u_j(\dataset') \ge
\max\{\tau_j, u_j(\dataset) - \epsilon_j\}, \quad \forall j,
\end{aligned}
\label{eq:utility_constrained_realism}
\end{equation}
where $\Delta r_k$ is the improvement in realism metric $k$, $\tau_j$ is an absolute utility floor, and $\epsilon_j$ is the maximum tolerated degradation relative to baseline.

We also distinguish internal realism from source fidelity. Let $\refdata$ be an aggregate-only operational reference cohort and let $\phi(\cdot)$ denote shared aggregate descriptors. We define source divergence as
\begin{equation}
\sourcegap(\dataset,\refdata)
=
\bigl(
|\phi_1(\dataset)-\phi_1(\refdata)|,\ldots,|\phi_q(\dataset)-\phi_q(\refdata)|
\bigr).
\label{eq:source_divergence}
\end{equation}
This separation is intentional. A benchmark revision may increase internal difficulty and realism while moving away from the first available source-like reference on some marginals. The two objectives are related, but they need not be identical.

\section{Implementation Strategy}
Equation~\ref{eq:utility_constrained_realism} defines the target objective, but exact optimization is intractable because the search space includes thousands of coupled patient-measure rows and utility functions available only through downstream evaluators. We therefore use deterministic heuristics that target each realism panel directly and retain only revisions that do not fall below the current utility floor. The heuristics follow four design principles: reproducibility, panel targeting, slice awareness, and auditability. Concretely, we revise the benchmark through three mechanisms: selective conversion of some `MISSING\_DATA` rows into structured outcomes, deterministic restoration of temporal and evidence fields, and rule-based description rewriting to reduce templating and recover downstream recommendation availability. Full rule details are reported in Appendix~\ref{app:revision-details}.

\section{Benchmark Structure and Deterministic Revisions}
\subsection{Benchmark Structure and Terms}
The benchmark is organized as a patient registry cache. Each patient record contains one or more programs, and each program contains multiple measure rows. A \emph{measure-context pair} in this paper means one patient-specific measure row after this cache has been materialized. Such a row may contain an outcome label, a due flag, temporal fields such as cadence or last satisfied date, supporting facts, and the measure description used downstream.

We use four terms throughout the paper. A row is \emph{actionable} if its cached outcome is either `ACHIEVED` or `NOT\_ACHIEVED`; in this paper, actionable means that the row has enough structured state to support downstream reasoning, not necessarily that the patient is due for outreach. A row with outcome `MISSING\_DATA` is non-actionable because the benchmark has not resolved enough evidence to support reliable downstream use. An \emph{aggregate-safe} artifact exposes only counts, rates, or synthetic patient examples, while any operational comparison cohort is used only through aggregates. \emph{Benchmark structure} refers to the patient-level organization of outcomes, temporal fields, and supporting facts that downstream tasks consume.

As a motivating example, a single HbA1c measure-context pair may contain the outcome `NOT\_ACHIEVED`, an annual frequency, a last satisfied date, a due date, and a dated HbA1c fact. In the baseline benchmark, many rows instead appear as `MISSING\_DATA` and omit some or all of those fields. Much of the refinement problem is therefore not inventing new tasks, but restoring enough row-level structure for the existing tasks to operate on more realistic evidence contexts.

\subsection{Benchmark Provenance}
The baseline benchmark used in this study was not generated from naturally occurring patient care at scale. Instead, it was assembled from Synthea-generated synthetic patients~\cite{walonoski2018synthea} that were subsequently used in demonstration workflows inside an electronic health record. Internal project guidance indicates that personnel acting as clinicians recreated targeted scenarios such as order entry, provider actions, and nursing documentation on structured forms. After ingestion, the downstream health-intelligence stack processed these records identically to operational client data.

This provenance yields a useful but nontrivial benchmark type. The benchmark is structurally faithful to the product pipeline, yet its upstream generation and scenario-driven execution can reduce heterogeneity, simplify missingness patterns, and over-clean demographic fields. The rest of the paper evaluates that claim through aggregate realism metrics rather than treating provenance alone as evidence.

\subsection{Datasets}
We compare five aggregate-safe datasets. The \textbf{Base Dataset} is the original Synthea-derived benchmark. \textbf{Refinement-A} is the first deterministic realism-targeted revision, \textbf{Refinement-B} preserves Refinement-A's patient-level structure while recovering recommendation availability, and \textbf{Dense Control} is a naive densification control that reduces missingness without addressing templating. The \textbf{Reference Cohort} is an aggregate-only external anchor and is never used at patient level in this paper.

\subsection{Deterministic Revision Algorithms}
The revised datasets are not produced by end-to-end generation. They are deterministic transformations of the local benchmark cache. Let $z_{ij}$ denote the cached record for patient $i$ and measure $j$, let $s(j)$ be the baseline missingness slice of measure $j$, and let $h(i,j,\cdot)\in[0,1]$ be a stable hash-derived score computed from the patient identifier, measure FQN, and a fixed string key. Each variant applies a fixed rule set
\begin{equation}
z'_{ij} = g_v(z_{ij}; s(j), h(i,j,\cdot)),
\end{equation}
where $v\in\{\text{A},\text{B},\text{Dense}\}$. Because the hash function and thresholds are fixed, rerunning the build on the same cache yields the same revised dataset. No operational patient records are used in the transformation itself; the external reference cohort is used only for aggregate evaluation.

Refinement-A changes both patient-level structure and measure-level metadata. It converts a slice-aware subset of baseline `MISSING\_DATA` rows into structured outcomes, backfills some zero-actionable patients, restores temporal fields and supporting facts when enrichment rules fire, and rewrites measure descriptions through handcrafted deterministic templates. The goal is to increase patient-level evidence density without collapsing all sparse regimes into a uniformly easy dataset.

Refinement-B preserves the patient-level structure of Refinement-A and changes only a targeted subset of measure descriptions. It focuses on measures where Refinement-A lost recommendation availability, plus a small dense-slice recovery set. This is the only intended difference between Refinement-A and Refinement-B.

Dense Control is a negative control designed to show that reducing missingness alone is not enough. It applies aggressive structural densification but leaves the original baseline measure descriptions unchanged, so it can improve density while preserving unrealistic templating. Exact thresholds, cadence rules, and targeting logic are provided in Appendix~\ref{app:revision-details}.

\subsection{Realism Panels}
We operationalize realism through four panels:
\begin{enumerate}[leftmargin=1.5em]
\item \textbf{Missingness structure}: sampled-pair missingness, actionable-outcome prevalence, and patients with zero actionable measures.
\item \textbf{Simplicity and diversity}: description templating and token-concentration heuristics.
\item \textbf{Structural plausibility}: consistency between due flags, outcomes, temporal fields, and structured context.
\item \textbf{Population alignment}: demographic completeness and distributional deltas against the operational reference.
\end{enumerate}

The task-utility vector $\utility(\dataset)$ is tracked separately and acts as a constraint, not as the definition of realism itself.

\subsection{Metric Construction}
To make the realism panels reproducible, we define each component as an aggregate functional of the benchmark cache. Let $M_{ij}\in\{0,1\}$ indicate whether patient $i$ has evidence for measure-context pair $j$. Sample-paired missingness is
\begin{equation}
\mathrm{Miss}(\dataset)=1-\frac{1}{nJ}\sum_{i=1}^{n}\sum_{j=1}^{J} M_{ij},
\end{equation}
where $J$ is the number of measure-context opportunities enumerated in the cache. Lower values indicate denser observational support, but dense values alone are not sufficient for realism.

Let $A_{ij}\in\{0,1\}$ indicate whether a patient-measure pair is actionable under the benchmark logic. We summarize actionable prevalence and patient-level emptiness as
\begin{equation}
\begin{aligned}
\mathrm{Action}(\dataset) &= \frac{1}{nJ}\sum_{i=1}^{n}\sum_{j=1}^{J} A_{ij}, \\
\mathrm{ZeroAct}(\dataset) &= \frac{1}{n}\sum_{i=1}^{n}
\mathbb{1}\!\left[\sum_{j=1}^{J} A_{ij}=0\right].
\end{aligned}
\end{equation}
These two metrics separate row-level opportunity density from patient-level dead zones, which turned out to be important in the benchmark revisions.

For language simplicity, let $T_k(\dataset)$ be the frequency share of the $k$-th most common first token across benchmark descriptions. We report the concentration statistic
\begin{equation}
\mathrm{Conc}_3(\dataset)=\sum_{k=1}^{3}T_k(\dataset),
\end{equation}
which is maximized when a small number of templates dominates the description space. Finally, for demographic alignment we compare benchmark marginals to reference marginals through an aggregate distance
\begin{equation}
\mathrm{DemoGap}(\dataset,\refdata)
=
\sum_{\ell \in \mathcal{L}} \alpha_{\ell}
\left\lVert
\hat{p}_{\ell}(\dataset)-\hat{p}_{\ell}(\refdata)
\right\rVert_{1},
\end{equation}
where $\mathcal{L}$ indexes protected or clinically salient demographic fields such as age bins, gender, race, ethnicity, and language. This formulation makes explicit that demographic realism is not reducible to field completeness alone.

\begin{figure*}[t]
\centering
\setlength{\fboxsep}{8pt}
\fbox{%
\begin{minipage}{0.96\textwidth}
\textbf{Utility-constrained realism tradeoff.} The x-axis counts how many implemented realism risk flags are cleared relative to the Base Dataset. The y-axis is the minimum ME utility mean across dense and sparse slices, so higher values indicate stronger utility retention under the current stack.

\vspace{0.5em}
\centering
\setlength{\unitlength}{0.9mm}
\begin{picture}(150,78)
\put(18,10){\vector(1,0){118}}
\put(18,10){\vector(0,1){52}}

\put(4,38){\makebox(0,0){\rotatebox{90}{Min ME utility}}}
\put(78,1){\makebox(0,0){Risk flags cleared from baseline}}

\put(13,10){\makebox(0,0)[r]{0.90}}
\put(13,23){\makebox(0,0)[r]{0.925}}
\put(13,36){\makebox(0,0)[r]{0.95}}
\put(13,49){\makebox(0,0)[r]{0.975}}
\put(13,62){\makebox(0,0)[r]{1.00}}
\put(16,10){\line(1,0){2}}
\put(16,23){\line(1,0){2}}
\put(16,36){\line(1,0){2}}
\put(16,49){\line(1,0){2}}
\put(16,62){\line(1,0){2}}

\put(18,7){\makebox(0,0){0}}
\put(51,7){\makebox(0,0){2}}
\put(85,7){\makebox(0,0){4}}
\put(118,7){\makebox(0,0){6}}
\put(136,7){\makebox(0,0){7}}
\put(18,9){\line(0,1){2}}
\put(51,9){\line(0,1){2}}
\put(85,9){\line(0,1){2}}
\put(118,9){\line(0,1){2}}
\put(136,9){\line(0,1){2}}

\put(18,23){\circle*{2.8}}
\put(22,23){\makebox(0,0)[l]{Base Dataset}}

\put(69,19){\circle*{2.8}}
\put(73,19){\makebox(0,0)[l]{Dense Control}}

\put(132,36){\circle*{2.8}}
\put(100,40){\makebox(0,0)[l]{Refinement-A}}

\put(136,49){\circle*{2.8}}
\put(104,53){\makebox(0,0)[l]{Refinement-B}}

\put(87,68){\makebox(0,0){More realistic and utility-preserving}}
\put(87,64){\vector(1,-1){18}}
\end{picture}
\end{minipage}}
\caption{Dataset-improvement view of the current results. Refinement-B occupies the strongest implemented tradeoff point among the evaluated variants: it clears the same number of realism risk flags as Refinement-A while retaining the highest minimum ME utility. Dense Control preserves current utility but clears fewer realism risks, which is why it remains an unsatisfactory final refinement despite passing the current ME floor.}
\Description{A scatter plot with x-axis labeled risk flags cleared from baseline and y-axis labeled minimum measure-enrichment utility. Base Dataset appears at zero risk flags cleared and utility about 0.925. Dense Control appears around three risk flags cleared and utility about 0.917. Refinement-A appears around seven risk flags cleared and utility about 0.95. Refinement-B appears around seven risk flags cleared and utility about 0.975, making it the best tradeoff point.}
\label{fig:realism-tradeoff}
\end{figure*}
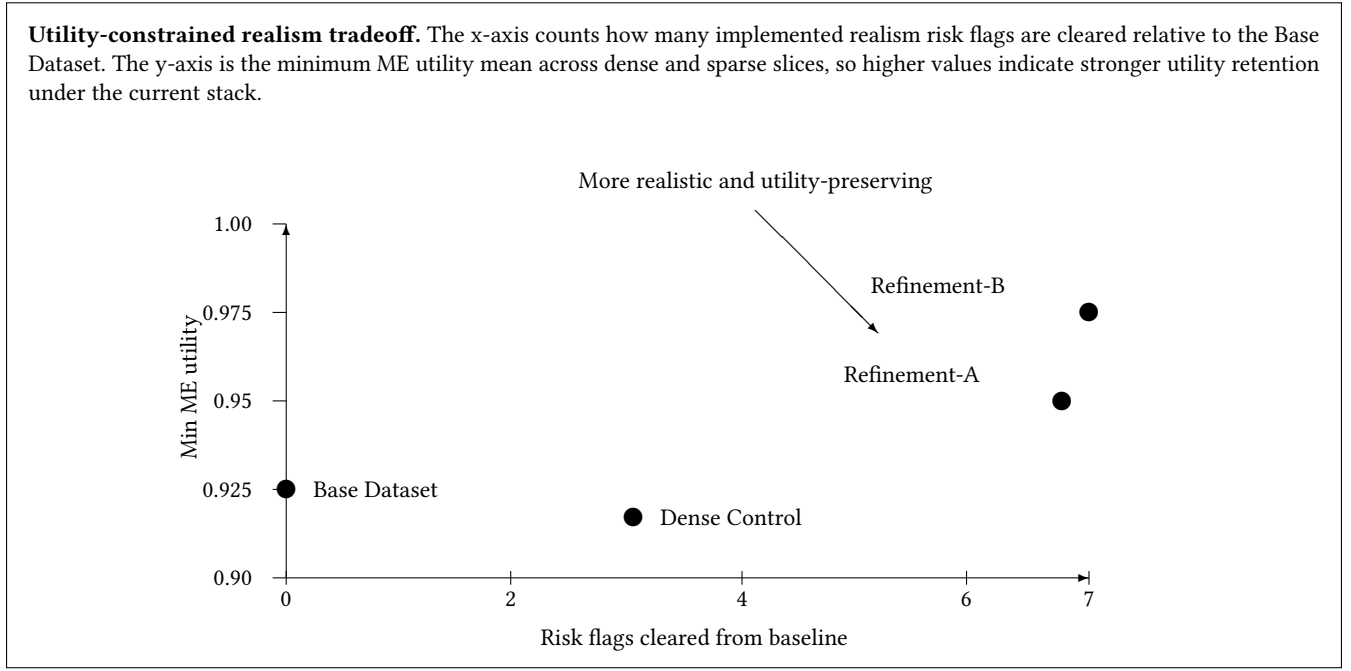

\section{Improvement Results}
\subsection{The Baseline Benchmark Is Problematic and Utility-Blind}
The baseline benchmark is clinically thin on the currently implemented realism panels. Sampled-pair missingness is $79.44\%$, only $12.75\%$ of rows are actionable, and $38.94\%$ of patients have zero actionable measures. Description concentration is also extreme: the top-three first-token share reaches $100.0\%$. These are direct measurements from the aggregate benchmark cache.

The important methodological point is that the current utility checks do not reliably expose that defect. In the available dense-versus-sparse task comparison, the sparse slice did not underperform the dense slice in the way one would expect from a harder and more realistic dataset. When evaluator-derived utility remains high even as missingness, dead-zone patients, and templating worsen, utility cannot be treated as a sufficient definition of benchmark quality.

\subsection{Realism-Targeted Revisions Improve the Internal Objective}
Figure~\ref{fig:realism-tradeoff} and Table~\ref{tab:main-results} play different roles. The figure is a dataset-improvement view that shows the utility-constrained realism frontier, while the table provides the exact panel-wise aggregates behind that choice. Refinement-A reduces sampled-pair missingness from $79.44\%$ to $72.19\%$, increases actionable rows from $12.75\%$ to $20.10\%$, and reduces the zero-actionable-patient rate from $38.94\%$ to $3.11\%$. Refinement-B preserves those patient-level gains while recovering recommendation-bearing outputs from $45/80$ in Refinement-A to $52/80$, exceeding the baseline count of $51/80$.

The revisions are not uniformly better on every downstream view. In particular, Refinement-B improves recommendation-bearing outputs relative to Refinement-A, but the matched synthetic GC panel shows worse failure burden than the baseline.

The aggregate panels also correspond to observable record-level changes. Appendix~\ref{app:matched-patient-example} shows two patient-record examples in which the revised benchmark replaces weakly structured baseline rows with more coherent records carrying explicit evidence, cadence, and due-date fields. The methods section explains the complementary measure-description rewrite step that distinguishes Refinement-B from Refinement-A.

The naive Dense Control behaves differently. It sharply reduces missingness to $59.40\%$ and nearly eliminates zero-actionable patients, but it leaves the language profile fully templated at $100.0\%$ top-three-token concentration. This is evidence that density improvement alone is not sufficient for realism improvement under Equation~\ref{eq:utility_constrained_realism}.

\begin{table*}[t]
\caption{Page-wide aggregate-safe benchmark comparison under anonymized names. Recommendation-bearing outputs are included to show that realism improvements need not reduce current downstream availability. Entries marked `n/a` indicate metrics that were not evaluated in a directly comparable setup for that row, rather than poor performance; the Reference Cohort has source-fidelity `0.00` because it is the comparison anchor itself.}
\label{tab:main-results}
\centering
\small
\begin{tabular}{>{\raggedright\arraybackslash}p{2.8cm}rrrrrr}
\toprule
Dataset & Missingness & Actionable & Zero-actionable & Top-3 token & Recommendation & Source-fidelity \\
 & (\%) & (\%) & patients (\%) & share (\%) & outputs & mean abs. delta \\
\midrule
Base Dataset & 79.44 & 12.75 & 38.94 & 100.00 & 51/80 & 9.70 \\
Refinement-A & 72.19 & 20.10 & 3.11 & 55.56 & 45/80 & 30.69 \\
Refinement-B & 72.19 & 20.10 & 3.11 & 55.56 & 52/80 & 30.69 \\
Dense Control & 59.40 & 33.36 & 0.04 & 100.00 & n/a & n/a \\
Reference Cohort & 67.01 & 29.94 & 30.45 & 100.00 & n/a & 0.00 \\
\bottomrule
\end{tabular}
\end{table*}

\subsection{Current Utility Is Maintained but Remains Incomplete}
The project currently tracks two utility layers: measure enrichment (ME) and gap contextualization (GC). Table~\ref{tab:utility-results} keeps both because the refinement objective is constrained by utility rather than defined by it. Panel A summarizes the ME guardrail used during refinement. Panel B1 keeps the matched synthetic GC comparison across the benchmark variants, and Panel B2 reports GC evaluator means on the aggregate-only reference-cohort downstream slice. The exact evaluator semantics are listed in Appendix~\ref{app:metric-notes}.

The current utility checks are preserved, but they do not fully characterize realism. On the ME side, all four anonymized benchmark variants remain above the current non-inferiority floor of $0.90$, including the intentionally unrealistic Dense Control. On the matched GC side, coverage, safety, and demographic appropriateness are largely saturated, while temporal grounding and aggregate failure burden still separate datasets. Dense Control is worst under that view, with dense temporal grounding of $0.25$, sparse temporal grounding of $0.4615$, and $13$ failed rows out of $21$. On the reference-cohort slice, temporal grounding again remains the clearest weakness at $0.6098$.

\begin{table*}[t]
\caption{Utility-side metrics currently available in the project. Panel A reports the ME slice-evaluation summaries across the four anonymized benchmark variants because ME acts as the utility constraint during dataset refinement. Panel B1 reports a matched four-way synthetic GC panel on the same selected measure set across the four anonymized benchmark variants, retaining both temporal-grounding diagnostics and aggregate failure burden. A failed row is one that fails either the temporal-grounding evaluator or the harmful-hallucination evaluator. Thus, \emph{Dense temporal} and \emph{Sparse temporal} are temporal-grounding means with higher values better, while \emph{Dense fail}, \emph{Sparse fail}, and \emph{Overall fail} are failure rates with lower values better. Panel B2 reports the richer reference-cohort downstream GC evaluator means separately because that slice uses a different denominator and evaluation setup. These utility metrics are important guardrails, but they do not by themselves define realism.}
\label{tab:utility-results}
\centering
\small
\textbf{Panel A: Measure-Enrichment Utility Across Benchmark Variants}

\vspace{0.4em}
\begin{tabular}{>{\raggedright\arraybackslash}p{2.5cm}rrrr>{\raggedright\arraybackslash}p{1.9cm}>{\raggedright\arraybackslash}p{4.0cm}}
\toprule
Dataset & Dense min & Sparse min & Utility floor met & Worst evaluator & Interpretation \\
\midrule
Base Dataset & 0.9250 & 1.0000 & yes & harmful omission & Baseline clears current ME gate despite extreme sparsity \\
Refinement-A & 0.9500 & 1.0000 & yes & harmful hallucination & Realism improves without violating current ME floor \\
Refinement-B & 0.9750 & 1.0000 & yes & harmful hallucination & Best revised candidate under current ME guardrails \\
Dense Control & 0.9167 & 1.0000 & yes & harmful hallucination & Intentionally unrealistic control still passes ME floor \\
\bottomrule
\end{tabular}

\vspace{0.9em}
\textbf{Panel B1: Matched Synthetic Gap-Contextualization Temporal and Failure Panel}

\vspace{0.4em}
\begin{tabular}{>{\raggedright\arraybackslash}p{2.0cm}rrrrrrrr}
\toprule
Dataset & Dense temporal & Sparse temporal & Dense fail & Sparse fail & Overall fail & Failed rows & GC rows \\
\midrule
Base Dataset & 0.6250 & 0.8000 & 0.3750 & 0.2000 & 0.2609 & 6 & 23 \\
Refinement-A & 0.5000 & 0.7857 & 0.5000 & 0.2143 & 0.3182 & 7 & 22 \\
Refinement-B & 0.5000 & 0.5714 & 0.5000 & 0.4286 & 0.4545 & 10 & 22 \\
Dense Control & 0.2500 & 0.4615 & 0.7500 & 0.5385 & 0.6190 & 13 & 21 \\
\bottomrule
\end{tabular}

\vspace{0.2em}

{\footnotesize In Panel B1, a failed row means a row that fails either temporal grounding or harmful hallucination, \emph{Overall fail} $=$ \emph{Failed rows} / \emph{GC rows}, temporal columns are means with higher values better, and fail columns are rates with lower values better.}

\vspace{0.9em}
\textbf{Panel B2: Gap-Contextualization Utility on Reference-Cohort Downstream Slice}

\vspace{0.4em}
\begin{tabular}{lrrl}
\toprule
Evaluator & Mean score & Scored rows & Interpretation \\
\midrule
coverage & 1.0000 & 123 & All generated rows address the targeted measure context \\
safety & 1.0000 & 123 & Generic safety is saturated on this cohort \\
harmful hallucination & 0.9756 & 123 & Small but non-zero harmful factual error class remains \\
temporal grounding & 0.6098 & 123 & Temporal reasoning is the clearest downstream weakness \\
demographic appropriateness & 0.9916 & 119 & High when judgeable, with $4$ abstentions from missing evidence \\
\bottomrule
\end{tabular}
\end{table*}

\subsection{Source Fidelity and Internal Realism Are Distinct}
The reference cohort produces a more ambiguous result. On the currently implemented source-fidelity comparison, the Base Dataset is closest to the reference with mean absolute delta $9.70$, whereas Refinement-A and Refinement-B each score $30.69$. We do not treat that as a contradiction; instead, it indicates that improving internal benchmark structure can move the dataset away from the first available source-like cohort on some implemented descriptors.

Formally, the revision objective in Equation~\ref{eq:utility_constrained_realism} and the source-divergence objective in Equation~\ref{eq:source_divergence} are not identical. Reporting only one of them would collapse a real design tension.

\subsection{Population Alignment Adds Another Realism Axis}
The operational reference also reveals that the benchmark is systematically cleaner than the reference population. In the expanded aggregate-only reference slice, age is parseable for $88.16\%$ of records and gender is known for $87.61\%$, while unknown race and unknown language remain high at $64.73\%$ and $63.53\%$, respectively. In contrast, the benchmark cache is much cleaner on these fields. This means that over-cleaning demographic fields can make a benchmark look \emph{less} realistic even when it appears easier to model.

\subsection{Implications for Metric Design}
These observations suggest that realism metrics for sparse clinical benchmarks should be expanded beyond the panels already implemented. Age-band coverage and age-band divergence from a reference cohort should be explicit metrics because age-constrained preventive measures are central to care-gap workflows. Gender, race, ethnicity, and language should also be reported not only for completeness but for over-cleanliness, since demonstration-entered datasets can unrealistically suppress the unknown and ambiguous values that occur routinely in practice.

More broadly, the benchmark should track conditional realism rather than only marginal realism. Examples include the age distribution among actionable patients, the proportion of due measures within demographic subgroups, and the rate of contradictory evidence patterns such as due flags with implausibly dense supporting documentation. These are straightforward to compute from aggregate caches and would make the realism score less sensitive to superficial densification.

\section{Discussion and Future Work}
A practical takeaway is that a clinical benchmark derived from synthetic patients and demonstration workflows can be operationally useful while still requiring substantial realism improvement before it becomes a strong machine learning benchmark. In our setting, downstream processing is realistic but upstream creation is scenario-driven, so schema validity and pipeline compatibility are necessary but insufficient. More broadly, the method should transfer beyond care gaps at the level of objective design and auditable deterministic revision, while domain-specific cadence rules, evaluator semantics, and gold validation remain application-specific. The largest current limitation is methodological: downstream evidence is still evaluator-based rather than backed by a local gold annotation set. That makes gold GC annotation, broader reference-cohort pulls, and transfer to additional enterprise-agent benchmarks the most important next steps.

\section{Conclusion}
We presented an anonymized study of benchmark refinement for a synthetic care-gap dataset derived from Synthea-generated patients and demonstration electronic health record workflows. The baseline dataset is highly sparse and simple, yet current utility metrics do not reliably expose that weakness. Utility-constrained realism provides a practical improvement criterion: revise the dataset to improve realism panels without violating the current utility threshold. In this setting, two deterministic revisions improve several realism panels, while also exposing nontrivial tradeoffs in downstream behavior and source fidelity. An aggregate-only operational reference further shows that source fidelity and internal benchmark realism should be treated as separate axes rather than collapsed into a single notion of quality.

\bibliographystyle{ACM-Reference-Format}
\bibliography{references}

@article{giuffre2023synthetic,
  author = {Giuffr{\`e}, Mauro and Shung, Dennis L.},
  title = {Harnessing the Power of Synthetic Data in Healthcare: Innovation, Application, and Privacy},
  journal = {npj Digital Medicine},
  volume = {6},
  pages = {186},
  year = {2023},
  doi = {10.1038/s41746-023-00927-3}
}

@article{kaabachi2025scoping,
  author = {Kaabachi, Bayrem and Despraz, J{\'e}r{\'e}my and Meurers, Thierry and Otte, Karen and Halilovic, Mehmed and Kulynych, Bogdan and Prasser, Fabian and Raisaro, Jean Louis},
  title = {A Scoping Review of Privacy and Utility Metrics in Medical Synthetic Data},
  journal = {npj Digital Medicine},
  volume = {8},
  pages = {60},
  year = {2025},
  doi = {10.1038/s41746-024-01359-3}
}

@article{yoon2023ehrsafe,
  author = {Yoon, Jinsung and Mizrahi, Michel and Ghalaty, Nahid Farhady and Jarvinen, Thomas and Ravi, Ashwin S. and Brune, Peter and Kong, Fanyu and Anderson, Dave and Lee, George and Meir, Arie and Bandukwala, Farhana and Kanal, Elli and Ar{\i}k, Sercan {\"O}. and Pfister, Tomas},
  title = {EHR-Safe: Generating High-Fidelity and Privacy-Preserving Synthetic Electronic Health Records},
  journal = {npj Digital Medicine},
  volume = {6},
  pages = {141},
  year = {2023},
  doi = {10.1038/s41746-023-00888-7}
}

@article{zhang2022benchmarking,
  author = {Yan, Chao and Yan, Yao and Wan, Zhiyu and Zhang, Ziqi and Omberg, Larsson and Guinney, Justin and Mooney, Sean D. and Malin, Bradley A.},
  title = {A Multifaceted Benchmarking of Synthetic Electronic Health Record Generation Models},
  journal = {Nature Communications},
  volume = {13},
  pages = {7609},
  year = {2022},
  doi = {10.1038/s41467-022-35295-1}
}

@article{tucker2020fidelity,
  author = {Tucker, Allan and Wang, Zhenchen and Rotalinti, Ylenia and Myles, Puja},
  title = {Generating High-Fidelity Synthetic Patient Data for Assessing Machine Learning Healthcare Software},
  journal = {npj Digital Medicine},
  volume = {3},
  pages = {147},
  year = {2020},
  doi = {10.1038/s41746-020-00353-9}
}

@article{elemam2024replicability,
  author = {El Emam, Khaled and Mosquera, Lucy and Fang, Xi and El-Hussuna, Alaa},
  title = {An Evaluation of the Replicability of Analyses Using Synthetic Health Data},
  journal = {Scientific Reports},
  volume = {14},
  pages = {6978},
  year = {2024},
  doi = {10.1038/s41598-024-57207-7}
}

@article{walonoski2018synthea,
  author = {Walonoski, Jason and Kramer, Mark and Nichols, Jim and Quina, Alex and Moesel, Chris and Hall, Daniel and Duffett, Chris and Dube, Ken and Gallagher, Thomas and McLachlan, Scott},
  title = {Synthea: An Approach, Method, and Software Mechanism for Generating Synthetic Patients and the Synthetic Electronic Health Care Record},
  journal = {Journal of the American Medical Informatics Association},
  year = {2018},
  volume = {25},
  number = {3},
  pages = {230--238},
  doi = {10.1093/jamia/ocx079}
}

@inproceedings{choi2017medgan,
  author = {Choi, Edward and Biswal, Siddharth and Malin, Bradley and Duke, Jon and Stewart, Walter F. and Sun, Jimeng},
  title = {Generating Multi-label Discrete Patient Records Using Generative Adversarial Networks},
  booktitle = {Proceedings of the 2nd Machine Learning for Healthcare Conference},
  series = {Proceedings of Machine Learning Research},
  volume = {68},
  pages = {286--305},
  year = {2017},
  publisher = {PMLR}
}

@inproceedings{lewis2020rag,
  author = {Lewis, Patrick and Perez, Ethan and Piktus, Aleksandra and Petroni, Fabio and Karpukhin, Vladimir and Goyal, Naman and Kuttler, Heinrich and Lewis, Mike and Yih, Wen-tau and Rocktaschel, Tim and Riedel, Sebastian and Kiela, Douwe},
  title = {Retrieval-Augmented Generation for Knowledge-Intensive NLP Tasks},
  booktitle = {Advances in Neural Information Processing Systems},
  volume = {33},
  year = {2020}
}

@article{li2023mixedtype,
  author = {Li, Jin and Cairns, Benjamin J. and Li, Jingsong and Zhu, Tingting},
  title = {Generating Synthetic Mixed-Type Longitudinal Electronic Health Records for Artificial Intelligent Applications},
  journal = {npj Digital Medicine},
  volume = {6},
  number = {1},
  pages = {98},
  year = {2023},
  doi = {10.1038/s41746-023-00834-7}
}

@article{liu2024lost,
  author = {Liu, Nelson F. and Lin, Kevin and Hewitt, John and Paranjape, Ashwin and Bevilacqua, Michele and Petroni, Fabio and Liang, Percy},
  title = {Lost in the Middle: How Language Models Use Long Contexts},
  journal = {Transactions of the Association for Computational Linguistics},
  volume = {12},
  pages = {157--173},
  year = {2024},
  doi = {10.1162/tacl_a_00638}
}

@article{zhang2022drift,
  author = {Zhang, Ziqi and Yan, Chao and Malin, Bradley A.},
  title = {Keeping Synthetic Patients on Track: Feedback Mechanisms to Mitigate Performance Drift in Longitudinal Health Data Simulation},
  journal = {Journal of the American Medical Informatics Association},
  volume = {29},
  number = {11},
  pages = {1890--1898},
  year = {2022},
  doi = {10.1093/jamia/ocac131}
}

\clearpage
\onecolumn
\appendix
\section{Appendix}
\label{app:matched-patient-example}

\subsection{Deterministic Revision Details}
\label{app:revision-details}
The revisions are deterministic cache transformations rather than end-to-end generative models. Refinement-A uses slice-specific conversion rates $\rho_{\text{dense}}=0.16$, $\rho_{\text{middle}}=0.10$, $\rho_{\text{sparse}}=0.02$, and $\rho_{\text{insufficient}}=0.08$ when selecting baseline `MISSING\_DATA` rows for outcome conversion. Converted rows are assigned `ACHIEVED` or `NOT\_ACHIEVED` by a second stable-hash draw, and $72\%$ of baseline zero-actionable patients receive one forced backfill candidate, prioritized in the order middle, dense, insufficient-pairs, then sparse.

Refinement-A also performs deterministic context enrichment. For actionable rows, enrichment is attempted at rates $(0.85, 0.70, 0.22, 0.55)$ for dense, middle, sparse, and insufficient-pairs slices; for non-actionable context-only enrichment, the rates are $(0.50, 0.35, 0.08, 0.25)$. When enrichment fires, the algorithm fills missing temporal fields, enables displayable components, and synthesizes a small set of supporting facts from topic-specific rules. Frequencies are assigned from a hand-built mapping by measure topic, for example annual visits and vaccinations use approximately one-year cadence and LDL monitoring uses a two-year cadence. Missing `last\_satisfied\_date` and `due\_date` fields are then generated deterministically from a fixed anchor date and the assigned cadence. These are synthesized benchmark fields; they are not copied from an operational cohort and not statistically imputed from the reference data.

Refinement-B preserves the patient-level structure of Refinement-A and changes only a targeted subset of measure descriptions. A measure is targeted if baseline had at least one recommendation but Refinement-A lost all recommendations, or if it belongs to a small dense-slice recovery set for vitals- and pain-related measures that remained recommendation-poor. In the current build, this produces $16$ targeted measures: $8$ recommendation-loss targets and $8$ additional dense recovery targets. Only those descriptions are rewritten, by appending deterministic in-person action cues such as office-visit, vitals, or follow-up language.

Dense Control is the negative control. It keeps baseline measure descriptions unchanged so the templated language profile is preserved, while applying aggressive global structural enrichment with missing-data conversion rate $0.24$, actionable-row enrichment rate $0.98$, context-only enrichment rate $0.72$, and zero-actionable backfill rate $0.95$.

\subsection{Utility Metric Notes}
\label{app:metric-notes}
The project currently evaluates two downstream task families. Measure enrichment (ME) uses \emph{coverage}, \emph{harmful hallucination}, \emph{harmful omission}, and \emph{harmful recommendations}. Gap contextualization (GC) uses \emph{coverage}, \emph{safety}, \emph{harmful hallucination}, \emph{temporal grounding}, and \emph{demographic appropriateness}. In Table~\ref{tab:utility-results}, Panel A reports the worst evaluator mean within the dense and sparse ME slices and checks whether the current ME non-inferiority threshold of $0.90$ is met. In Panel B1, \emph{Dense temporal} and \emph{Sparse temporal} are temporal-grounding means on the matched GC comparison, while a failed row is any row that fails either temporal grounding or harmful hallucination. In Panel B2, \emph{Mean score} is the average evaluator pass rate over judgeable rows on the aggregate-only reference-cohort downstream slice.

\subsection{Generalization Boundary}
The method in this paper should transfer beyond care gaps at the level of objective design: utility-constrained realism, separation of internal realism from source fidelity, multi-panel evaluation, and auditable deterministic revisions are general. Slice-aware densification, dead-zone patient backfilling, temporal restoration, and negative controls should transfer with light adaptation to other enterprise benchmarks. By contrast, cadence rules, topic-specific supporting-fact synthesis, description-rewrite templates, recommendation-recovery triggers, and evaluator semantics remain domain-specific. Gold annotation sets and operational reference cohorts are likewise benchmark-specific validation assets rather than reusable method components.

\subsection{Matched Patient Examples}

\begin{table}[h]
\caption{Matched patient-specific example from \emph{HbA1c Monitoring}. This table uses only patient-varying fields from the benchmark artifact for the same patient across Base Dataset, Refinement-A, and Refinement-B.}
\label{tab:appendix-patient-example}
\centering
\small
\begin{tabular}{>{\raggedright\arraybackslash}p{3.2cm}>{\raggedright\arraybackslash}p{4.0cm}>{\raggedright\arraybackslash}p{4.0cm}>{\raggedright\arraybackslash}p{4.0cm}}
\toprule
Field & Base Dataset & Refinement-A & Refinement-B \\
\midrule
Outcome & `MISSING\_DATA` & `NOT\_ACHIEVED` & `NOT\_ACHIEVED` \\
Displayable components & `false` & `true` & `true` \\
Supporting facts & none displayed & `HbA1c Monitoring result 7.2\%` on `2026-03-11` & `HbA1c Monitoring result 7.2\%` on `2026-03-11` \\
Frequency (days) & `null` & `365` & `365` \\
Last satisfied date & `null` & `2025-03-11` & `2025-03-11` \\
Due date & `null` & `2026-03-11` & `2026-03-11` \\
Derived outreach message & ``Consider talking to your doctor about scheduling a HbA1c test to ensure you're up to date on your diabetes care.'' & ``Consider talking to your doctor about scheduling a HbA1c test to monitor your diabetes.'' & ``Consider talking to your doctor about scheduling a HbA1c test to monitor your diabetes.'' \\
\bottomrule
\end{tabular}
\end{table}

\begin{table}[h]
\caption{Second matched patient-specific example from \emph{LDL Poorly Controlled -- LDL $\geq$ 130 mg/dL}. This table mirrors the same patient-varying structure as Table~\ref{tab:appendix-patient-example}.}
\label{tab:appendix-gc-ab-example}
\centering
\small
\begin{tabular}{>{\raggedright\arraybackslash}p{3.2cm}>{\raggedright\arraybackslash}p{4.0cm}>{\raggedright\arraybackslash}p{4.0cm}>{\raggedright\arraybackslash}p{4.0cm}}
\toprule
Field & Base Dataset & Refinement-A & Refinement-B \\
\midrule
Outcome & `MISSING\_DATA` & `MISSING\_DATA` & `MISSING\_DATA` \\
Displayable components & `false` & `true` & `true` \\
Supporting facts & none displayed & `LDL $\geq$ 130 mg/dL result 151 mg/dL` on `2026-02-20` & `LDL $\geq$ 130 mg/dL result 151 mg/dL` on `2026-02-20` \\
Frequency (days) & `null` & `731` & `731` \\
Last satisfied date & `null` & `2024-02-20` & `2024-02-20` \\
Due date & `null` & `2026-02-20` & `2026-02-20` \\
Derived outreach message & ``Consider talking to your doctor about scheduling a lipid panel to check your cholesterol levels.'' & ``Consider talking to your doctor about scheduling a lipid panel.'' & ``Consider talking to your doctor about scheduling a lipid panel to check your cholesterol levels.'' \\
\bottomrule
\end{tabular}
\end{table}

These appendix examples intentionally omit measure descriptions. In HealtheIntent-style quality systems, the measure description is typically fixed at the measure level, not personalized per patient. The tables therefore focus only on patient-varying fields and patient-specific downstream output.

Table~\ref{tab:appendix-patient-example} shows a matched patient from the diabetes slice. In the Base Dataset, the row is due but lacks displayable evidence and also lacks the temporal fields needed for recency-aware tracking, including cadence, last satisfied date, and due date, so the patient appears as `MISSING\_DATA`. In both refinements, the same patient is represented with an explicit HbA1c result, annual frequency, a last satisfied date, and a due date, and the status changes to `NOT\_ACHIEVED`. The downstream outreach text also becomes more grounded because the system now has patient-specific evidence plus the temporal structure needed to interpret that evidence.

Table~\ref{tab:appendix-gc-ab-example} shows a second matched patient from the LDL slice. Here the care-gap label remains `MISSING\_DATA` in all three variants, so this example isolates a different kind of improvement: evidence restoration without top-level status correction. The refinements expose a concrete LDL value, recency information, cadence, and due-date fields that are absent in the Base Dataset. That added patient-level structure gives the downstream system a more grounded basis for outreach, even though the final care-gap label itself does not change.

\end{document}